\documentclass[conference]{IEEEtran}
\IEEEoverridecommandlockouts
\usepackage{cite}
\usepackage{amsmath,amssymb,amsfonts}
\usepackage{algorithmic}
\usepackage{textcomp}
\usepackage{xcolor}
\usepackage{color}
\usepackage{times}
\usepackage{epsfig}
\usepackage{graphicx}
\usepackage{amsmath}
\usepackage{amssymb}
\usepackage{verbatim}
\usepackage{graphicx}
\usepackage{subfig}
\usepackage{multirow}
\usepackage{array}
\usepackage{float}
\graphicspath{{images/}}
\definecolor{orange}{RGB}{255,127,0}
\def\BibTeX{{\rm B\kern-.05em{\sc i\kern-.025em b}\kern-.08em
    T\kern-.1667em\lower.7ex\hbox{E}\kern-.125emX}}
\begin{document}

\title{Predicting Group Cohesiveness in Images}

\author{\IEEEauthorblockN{ Shreya Ghosh \hspace{1cm} Abhinav Dhall }
\IEEEauthorblockA{
% \textit{\small Computer Science \& Engineering} \\
\textit{ Indian Institute of Technology Ropar}\\
\small Ropar, India \\
\tt\small {shreya.ghosh, abhinav}@iitrpr.ac.in}
\and
\IEEEauthorblockN{ Nicu Sebe}
\IEEEauthorblockA{
% \textit{ \small Information Engineering \& Computer Science} \\
\textit{University of Trento }\\
\small Trento, Italy \\
\tt\small sebe@disi.unitn.it}
\and
\IEEEauthorblockN{Tom Gedeon}
\IEEEauthorblockA{
% \textit{\small College of Engineering \& Computer Science} \\
\textit{ Australian National University}\\
\small Canberra, Australia \\
\tt\small tom@cs.anu.edu.au}

% \and

}

\maketitle

\begin{abstract}
The cohesiveness of a group is an essential indicator of the emotional state, structure and success of a group of people. We study the factors that influence the perception of group-level cohesion and propose methods for estimating the human-perceived cohesion on the group cohesiveness scale. In order to identify the visual cues (attributes) for cohesion, we conducted a user survey. Image analysis is performed at a group-level via a multi-task convolutional neural network. For analyzing the contribution of facial expressions of the group members for predicting the Group Cohesion Score (GCS), a capsule network is explored. We add GCS to the Group Affect database and propose the `GAF-Cohesion database'. The proposed model performs well on the database and is able to achieve near human-level performance in predicting a group's cohesion score. It is interesting to note that group cohesion as an attribute, when jointly trained for group-level emotion prediction, helps in increasing the performance for the later task. This suggests that group-level emotion and cohesion are correlated. 
\end{abstract}

% \begin{IEEEkeywords}
% component, formatting, style, styling, insert
% \end{IEEEkeywords}

\section{Introduction}
The concept of `teamwork' is defined as the collaborative effort of a group of people to accomplish a common goal in the most well-organized way \cite{salas2008teams}. One of the most important requirements for effective teamwork is cohesion. The main motivation of our work is to understand the human perception of Group Cohesiveness Score (GCS) \cite{treadwell2001group} from images and map the attributes to an Automatic Group Cohesion (AGC) pipeline. Group cohesiveness is defined as the measure of bonding between group members. Higher cohesiveness implies stronger group-level bonding. According to psychological studies, group cohesion depends on several factors such as members' similarity \cite{tajfel2010social}, group size \cite{carron1995group}, group success \cite{zaccaro1988effects} and external competition and threats \cite{thompson1981collaboration,rempel1997perceived}. The reason behind a strong group bonding can be positive (e.g. group success) or negative (e.g. threats). One of the key factors behind any group-level success is high group cohesiveness \cite{beal2003cohesion} as it affects group-level performance. Beal et al. \cite{beal2003cohesion} argue that group cohesion plays the most important role in group performance. 
% High group cohesiveness is one of the key factors behind any group-level success.
Similarly, group members' satisfaction \cite{hackman1992group} also plays an important role in deciding the cohesiveness of a group. %In fact, one personal level weakness can be overcome in a group.
\begin{figure}[t]
    \centering
     \subfloat{\includegraphics[width =1.7in,height=1.5in]{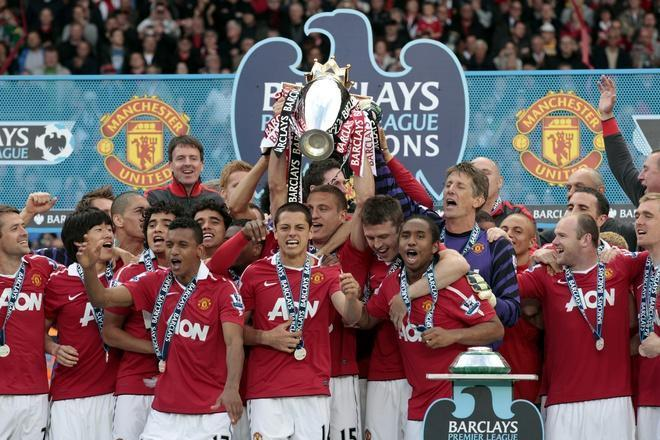}} \hspace{1mm}
    \subfloat{\includegraphics[width = 1.7in,height=1.5in]{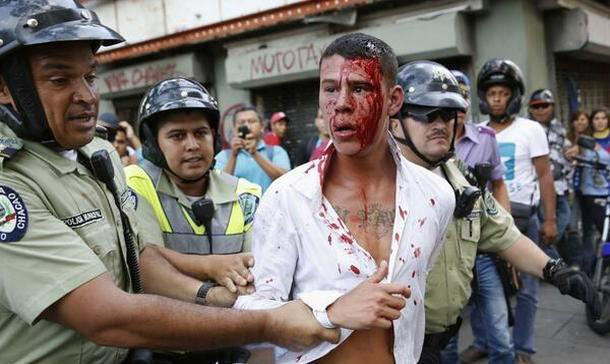}}
      \caption{\small The group of people in the left and the right images have high and low cohesion scores, respectively.}
      \label{prob}
      \vspace{-3mm}
\end{figure}
Hackman et al. \cite{hackman1992group} state that members belonging to a cohesive group have more satisfaction as compared to a non-cohesive group. Myers \cite{myers1961team} indicates that people belonging to a cohesive group are less prone to anxiety and tension. Lott et al. \cite{lott1965group} found that group cohesion helps improve individual members' learning processes. Inspired by the aforementioned studies, in this work we are interested in investigating the following research questions:

\begin{itemize}
\item \emph{How useful are holistic (image-level) and facial information for predicting cohesion in a group?}
\item \emph{What are the factors that affect the perception of the cohesiveness in a group?}
\item \emph{What is the usefulness of cohesiveness as an attribute for tasks such as group emotion prediction?}
\end{itemize}

In this work, we investigate AGC analysis from an early prediction perspective. This can also be viewed as a first impression of a group's cohesion, similar to the early personality assessment \cite{ponce2016chalearn} problem in affective computing. The main contributions of this paper are as follows:
\begin{enumerate}
    \item \textit{To the best of our knowledge, this is the first study proposing AGC prediction in images; }

    \item \textit{We compare two cohesion models, representing scene (holistic) and face-level information respectively, and show that the former contributes more to the perception of cohesion;}

    \item \textit{We label and extend the Group Affect Database \cite{dhall2017individual} with group cohesion labels and propose the \textbf{GAF Cohesion database} (sample images from the database are shown in Fig. \ref{prob}); }

    \item \textit{From our experimental results, we observed that the perceived group emotion is related to group cohesiveness (Section \ref{exp}).}
\end{enumerate}

The rest of the paper is structured as follows: Section \ref{sec:prior_cohesion} describes the prior works on Group cohesion. Section \ref{chal} explains the challenges involved in predicting the GCS task and the procedure of our survey. Section \ref{data} discusses the data and labeling process. The details of the proposed methods are described in Section \ref{pro}. Experiments are discussed in Section \ref{exp}. Section \ref{sec:visual} describes regarding the visual attributes that our network learn. Conclusion, limitations and future research directions are discussed in Section \ref{con}.

\section{Prior Work}
\label{sec:prior_cohesion}

\subsection{Group-level Cohesion (Psychological Aspects)}
According to Barsade et al. \cite{barsade1998group}, several factors impact the perception of a group's cohesion and emotion. The authors \cite{barsade1998group} argued that social norms and constraints (i.e. interpersonal bonding and individual emotional responses) are important cues for group emotion and cohesion. Gallagher et al. \cite{gallagher2009understanding} modelled the group as a min span tree based on facial locations and inferred the gender and age of group members using the group-level contextual information. Tajfel et al. \cite{tajfel2010social} stated that one of the main factors which affect a group's cohesiveness is its group members' similarity. Here, similarity can be measured in terms of their occupation, ethnicity, age and relationship etc. This may also imply that due to these factors group members may have a similar point of view about certain issues, which may cause strong bonding between them. Another interesting study by Carron et al. \cite{carron1995group} suggested that a small group implies strong cohesion. The reason behind this is that as the number of group members increases, their opinions may vary. This may lead to weaker cohesiveness as compared to small groups. Zaccaro et al. \cite{zaccaro1988effects} argued that group-level success (towards a task) is another factor, which influences cohesiveness, along with the group's size and its members' similarity. Apart from the positive factors, some negative factors may also influence a group's cohesiveness. Several studies \cite{thompson1981collaboration,rempel1997perceived} revealed that threats to a group and competition with another group may also increase a group's cohesiveness.

In a seminal work, Hung et al. \cite{hung2010estimating} studied group cohesion in a constrained environment using audiovisual-based group meeting data. Several audio and video features were extracted to test their importance on group cohesion. For audio analysis, pauses between individual turns, pauses between floor exchanges, turn lengths, overlapping speech, prosodic cues etc. are taken into consideration. Similarly, video features include pauses between individual turns, pauses between floor exchanges, motion turn lengths, overlapping visual activity, visual energy cues, ranking participants' features and group distribution features etc. Further, an SVM based classifier is used for predicting overall cohesion score. To the best of our knowledge, this is the first work that investigates the automatic cohesion of a group of people in videos.

\subsection{Study of `Group of People'}
In recent years, computer vision researchers have studied automatic analysis of `group of people' for different tasks. In an interesting work, Chang et al. \cite{chang2010group} predicted group-level activity via hierarchical agglomerative as well as the divisive clustering algorithm. In order to track the group-level activity, multiple cameras are placed in different environments (e.g. in an abandoned prison yard) which first detect group related information such as group formation, dispersion and distinct groups. Further, it investigates motion patterns (\emph{Loitering, Fast Moving, Approaching, Following}) and behaviour (e.g. Flanking, Agitation, Aggression). In another work, Wang et al. \cite{wang2015leveraging} proposed a method to infer the relationship between group members via geometric structure and appearance based features of the group. The AMIGOS database \cite{miranda2017amigos} has been recently proposed to study different aspects of affect in a group-level setting.

\subsection{Group Emotion}
One of the first group emotion analyses was proposed by Dhall et al. \cite{dhall2015automatic}. They proposed the Group Expression Model (GEM) to predict happiness intensity of a group of people in images. Several other studies \cite{MoodMeter,HuangDZGP15,li2016happiness,sun2016lstm,wei2017new,guo2017group,ghosh2018automatic} mainly extracted scene, face and pose features to predict group emotion. In another recent paper, Singh et al. \cite{singh2017individuals} studied the effect of a group on a person's smile. They evaluated the usefulness of visual features in predicting the task.

\begin{figure}[t]
    \centering
      \includegraphics[width=\linewidth]{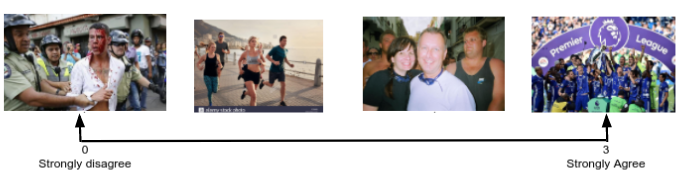}  
      \caption{\small The labels below the images above are based on the group cohesion scale (Treadwell et al. \cite{treadwell2001group}).}
      \label{scale} 
      \vspace{-5mm}
\end{figure}

\section{Challenges and Survey Result}
\label{chal}
\begin{figure}[t]
    \centering
      \includegraphics[width=\linewidth]{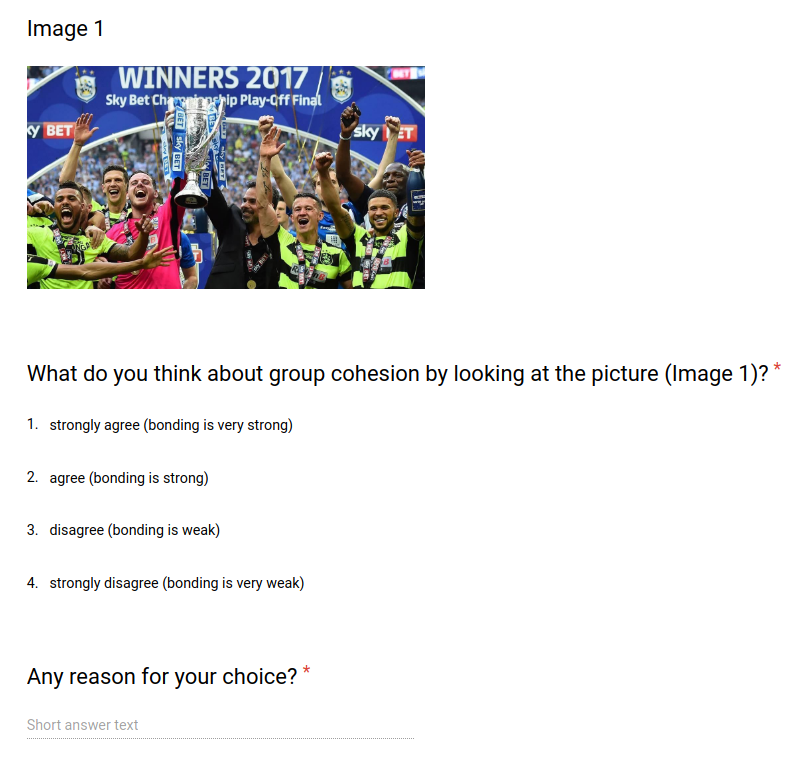}
       \caption{\small Screen shot of the user survey for understanding the factors, which effect the perception of a group's cohesiveness.}
      \label{survey}
      \vspace{-5mm}
\end{figure}

\begin{figure*}[!htbp]
\textbf{\hspace{18mm}Images \hspace{25mm} Word Cloud  \hspace{17mm} Survey Responses \hspace{7mm} Network Predictions}
\vspace{2mm}
\centering
\subfloat{\includegraphics[width = 2in,height=1.5in]{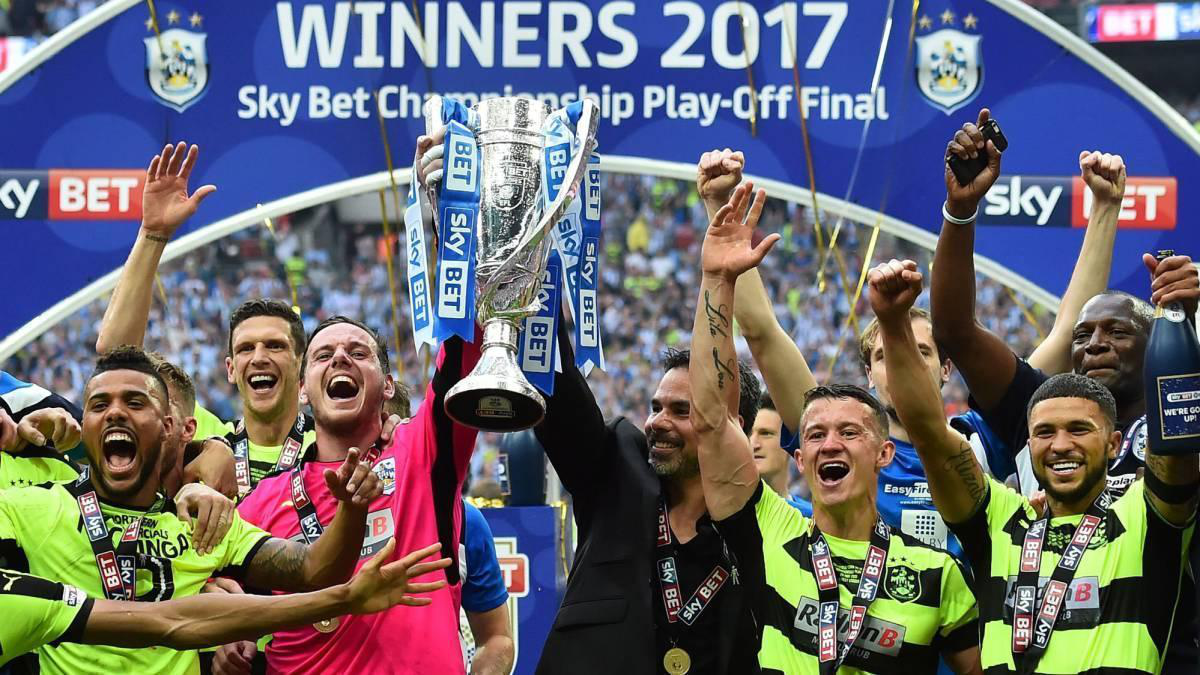}} \hspace{1mm}
\subfloat{\includegraphics[width = 1.5in,height=1.5in]{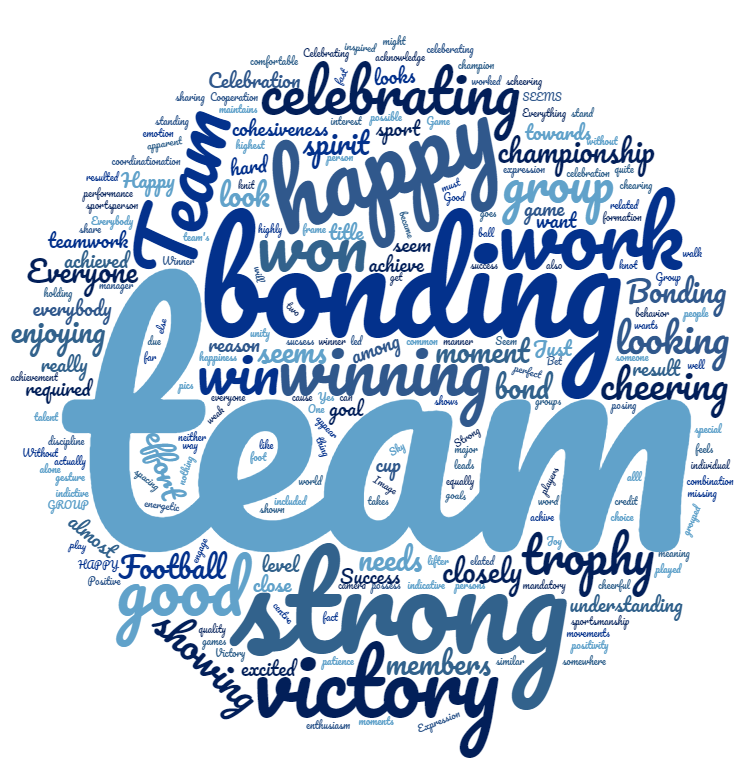}} \hspace{1mm}
\subfloat{\includegraphics[width = 1.5in,height=1.5in]{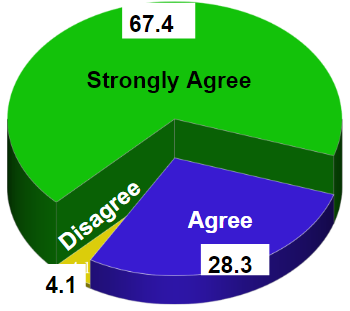}} \hspace{1mm}
\subfloat{\includegraphics[width = 1.3in,height=1.5in]{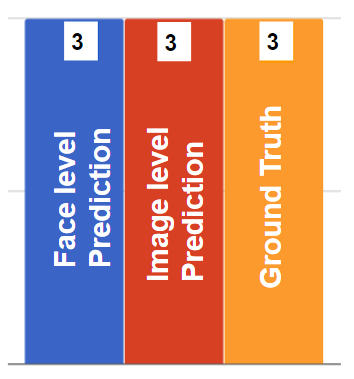}}
\\
\subfloat{\includegraphics[width = 2in,height=1.5in]{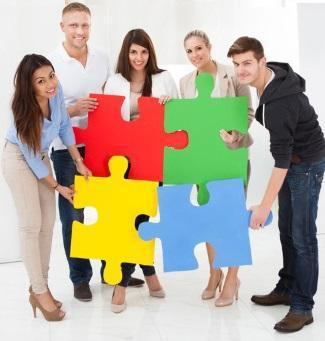}}\hspace{1mm}
\subfloat{\includegraphics[width = 1.5in,height=1.5in]{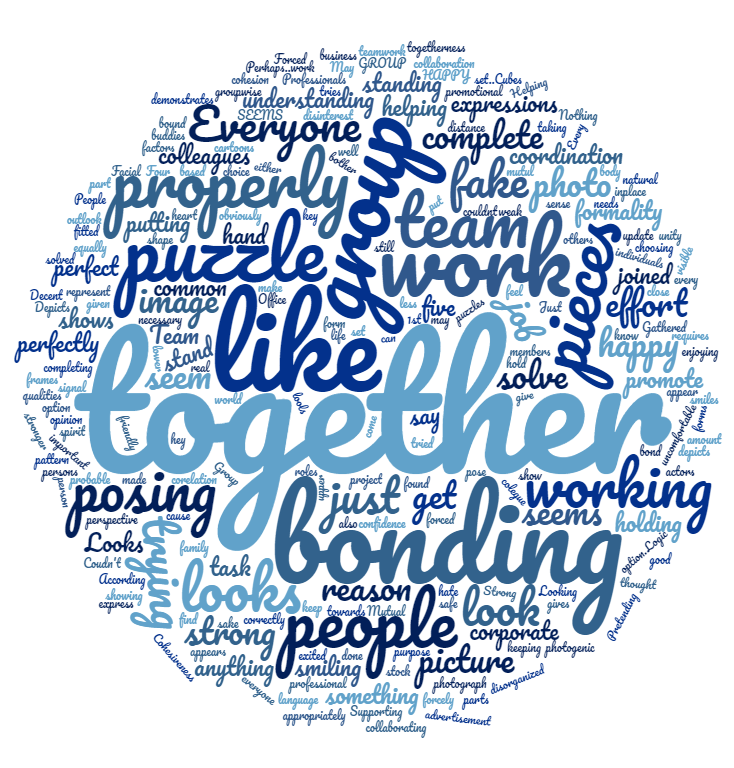}}\hspace{1mm}
\subfloat{\includegraphics[width = 1.5in,height=1.5in]{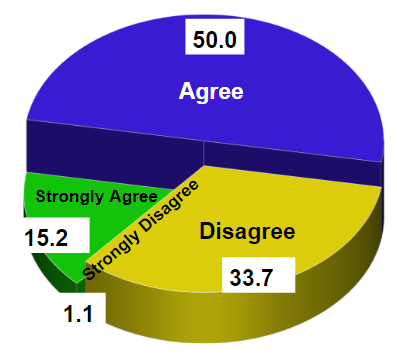}}\hspace{1mm}
\subfloat{\includegraphics[width = 1.3in,height=1.5in]{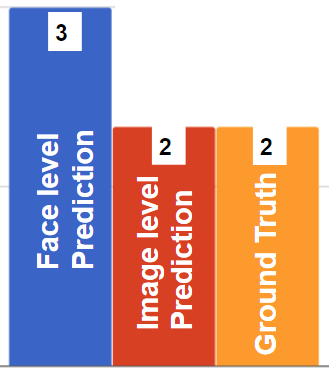}}
\\
\subfloat{\includegraphics[width = 2in,height=1.5in]{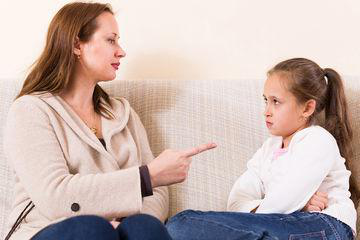}}\hspace{1mm}
\subfloat{\includegraphics[width = 1.5in,height=1.5in]{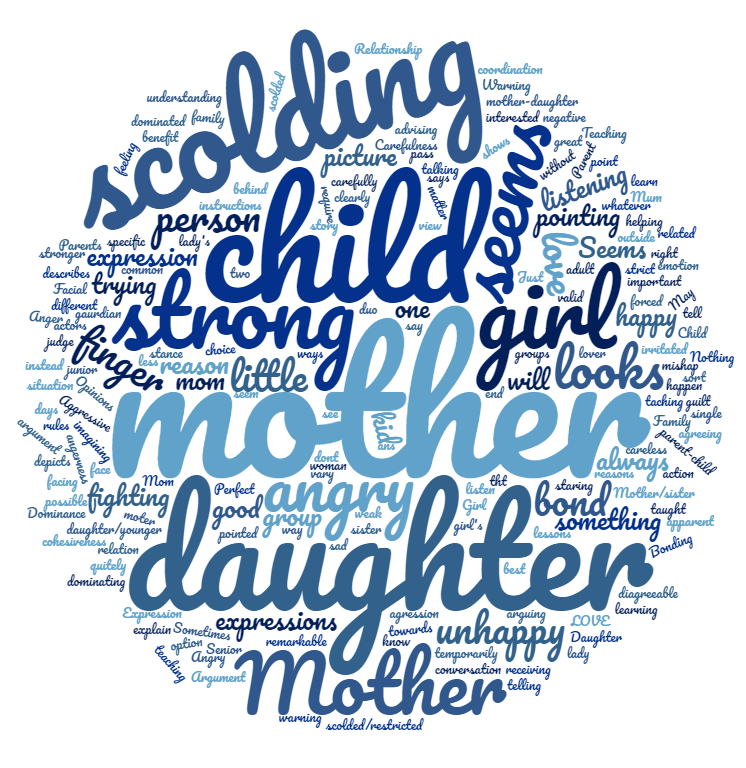}}\hspace{1mm}
\subfloat{\includegraphics[width = 1.5in,height=1.5in]{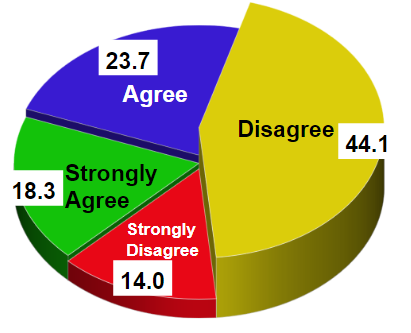}}\hspace{1mm}
\subfloat{\includegraphics[width = 1.3in,height=1.5in]{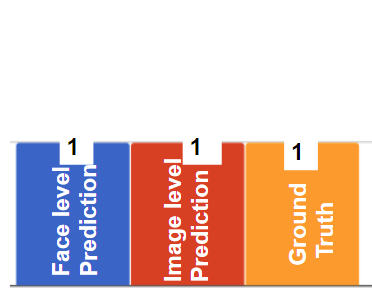}}
\\
\subfloat{\includegraphics[width = 2in,height=1.5in]{neg_91.png}}\hspace{1mm}
\subfloat{\includegraphics[width = 1.5in,height=1.5in]{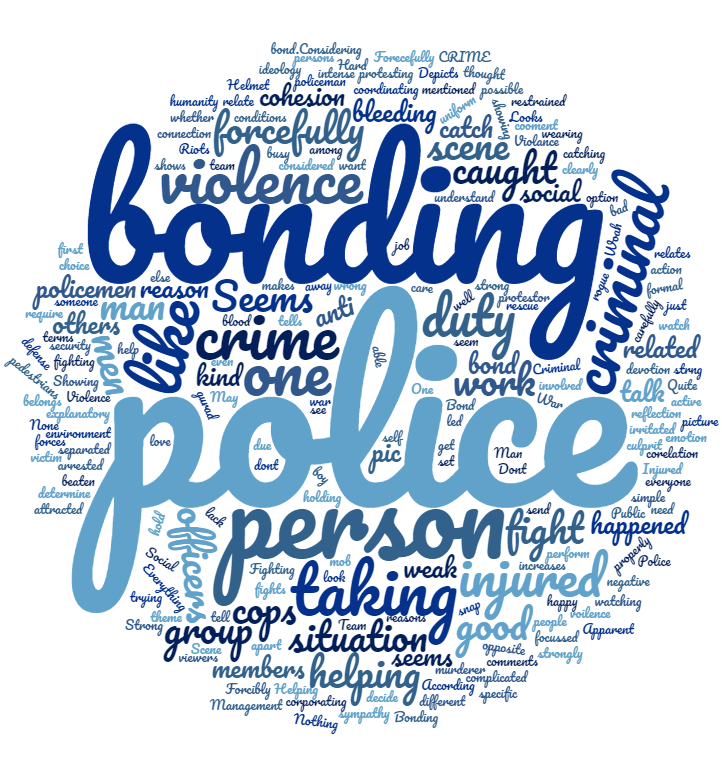}}\hspace{1mm}
\subfloat{\includegraphics[width = 1.5in,height=1.5in]{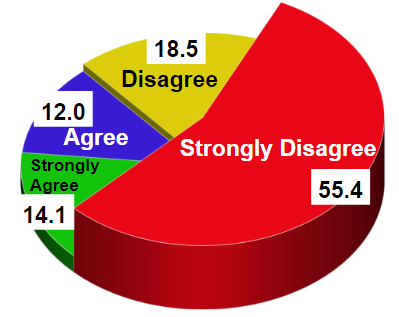}}\hspace{1mm}
\subfloat{\includegraphics[width = 1.3in,height=1.5in]{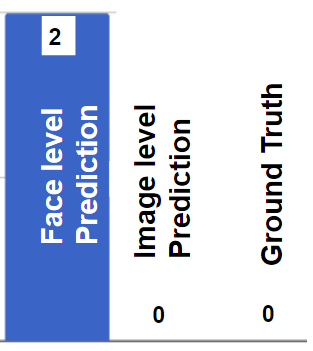}}

\caption{\small Survey results: The first column is the image. The second column represents the word cloud of keyword responses (responses against the reason field as shown in survey form Fig. \ref{survey}) and the third column consists of participant responses for a group's cohesion score. (Colour code for 3$^{rd}$ column: {\color{green}green}= strongly agree, {\color{blue}blue}= agree, {\color{yellow}yellow}= disagree and {\color{red}red}= strongly disagree) The fourth column shows the model prediction along with ground truth label for these images. For the 4$^{th}$ column {\color{blue}blue}= face-level prediction, {\color{red}red}= image-level prediction and {\color{orange}orange}= ground truth label). Prediction results are in the range [0 3]. In the results, the face-level network predicts the level of cohesion on the basis of emotion intensity similarity (e.g.  it detects smile faces across image 2 and thus it predicts it as high cohesion). Similarly, it can not predict correctly in case of 2$^{nd}$ and 4$^{th}$ image. [Best viewed in colour]}
\label{survey_result}
\vspace{-5mm}
\end{figure*}

This section describes the challenges involved in designing an AGC prediction network. To design an automated system for AGC prediction, we wish to know the factors which affect the perception of cohesion of a group. In the existing literature, the perception of members' similarity \cite{tajfel2010social} is claimed to be a vital visual cue; however, the first perception after viewing an image differs considerably from person to person. To understand the important visual cues, we have conducted a survey. The survey is conducted via Google form with 102 participants. There are 59 male and 43 female participants belonging to age group 22-54 years. The participants are from different backgrounds like student, businessman, corporate employee etc. The form consists of 24 images (as shown in Fig. \ref{survey}) of groups of people in different contexts and having different GCS values (6 images for each GCS value). Based on Treadwell et al. \cite{treadwell2001group}, we use four levels of cohesion. Before filling in the form, the participants are familiarized with the concept of group cohesion labels \cite{treadwell2001group} with images. The participants have to select one of the four cohesion levels for each image and they have to provide reasons behind their choice. Thus, we are provided some keywords related to the AGC score and corresponding image. After analyzing the responses, we get the statistics as shown in Fig. \ref{survey_result}. From the word clouds of Fig. \ref{survey_result}, we can see that `team', `bonding' and `together' are the most frequent keyword responses which indicate that we are dealing with group-level effects.
%The voting system indicates the `strong' cohesion between group members in the first row of Fig. \ref{survey_result}.
Further, `winning', `trophy', `work', `scolding', `fight' etc. reflect some holistic level features which motivate us to study image-level analysis. Similarly, some keywords such as `happy', `cheering', `angry', `violence' etc. tell about the mood of the individuals as well as the group. Thus, the survey motivates us to utilize both image-level features and face-level emotion features of an image. Our experiments are based on the understandings from the survey.
%Broadly, the findings from the survey can be divided into holistic and individual level attributes.

\section{Database}
\label{data}
To create the database, we have used and extended the images from the GAF 3.0 database \cite{dhall2017individual}. GAF 3.0 has been created via web crawling of various keywords related to social events (for example \emph{world cup winners, wedding, family, laughing club, birthday party, siblings, riot, protest} and \emph{violence} etc.). Images are added to GAF 3.0 to get a total of 14,175 images. We split the data into three parts: 9,815 images for training, 4,349 images for validation and 3011 images for testing purposes.

\subsection{Data Labeling} 

The GAF 3.0 database was labelled by 5 annotators (3 females and 2 males) of age group 21-30 years. In order to annotate data, the survey results assist about human perception regarding AGC. We have labelled each image for its cohesiveness in the range [0-3] \cite{treadwell2001group} as shown in Fig. \ref{scale}. Treadwell et al. \cite{treadwell2001group} argued that it is better to have these four `anchor points' (i.e., \emph{strongly agree, agree, disagree} and \emph{strongly disagree}) instead of having low to high scores. The low to high score scaling may vary perception-wise from person to person. Thus, these soft scaled `anchor points' are reliable. Along with GCS, GAF 3.0 database is also labelled with three group emotions (\emph{positive, negative and neutral}) across the valance axis. Before the annotation, the annotators are familiarized with the concepts of GCS labels \cite{treadwell2001group} with corresponding images.

\begin{figure}[t]
    \centering
      \includegraphics[width=\linewidth]{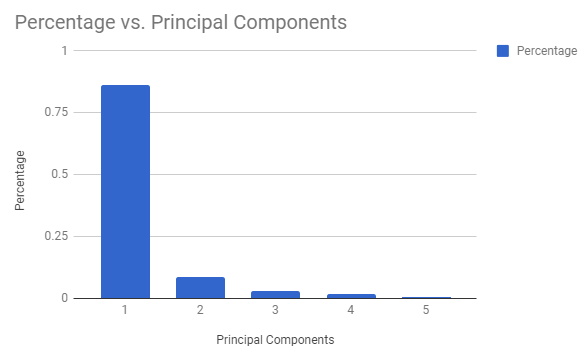} 
      \caption{\small The Figure shows the eigenvalues for the 5 principal components inter-rater variance. It is evident that the first principal component consists of 86\% of the distribution.}
      \label{eigen}
      \vspace{-5mm}
\end{figure}

\subsection{Annotation Statistics}

We further investigate the agreement between the annotators. The average variance and standard deviation between the annotators are 0.31 and 0.54, respectively. Further, we conduct a principal component analysis on the annotations as shown in Fig. \ref{eigen}. It is evident that approx. 86\% of the distribution lies in the first component, which suggests that there is a strong agreement between the annotators. Since the annotations were based on a `mutually exclusive category', we also measure the weighted generalized Cohen's kappa coefficient \cite{gwet2008computing} to determine the inter-rater agreement. The mean of the kappa coefficients value is 0.51. This also indicates high inter-rater agreeableness.

\section{Proposed Method}
\label{pro}
In this section, we discuss our approach for AGC analysis. We examined two networks one of which examines the image as a whole and another examines the facial expression of the group members.
\begin{table}[b]
\centering
\begin{tabular}{|c|c|c|c|}
\hline
\textbf{GAF 3.0} & \textbf{Ours} & \textbf{VGG 16} & \textbf{AlexNet} \\ \hline
MSE, GCS     & 0.8181         & 0.8967             & 1.0375                       \\ \hline
\begin{tabular}[c]{@{}l@{}}Accuracy(\%)\\ Group Emotion \end{tabular}    & 85.58         & 40.26             & 72.21                       \\ \hline
\end{tabular} \vspace{ 2mm}
\caption{\small GCS and emotion recognition comparison.}
\label{sota}
\vspace{-3mm}
\end{table}

\subsection{Image-level Analysis}
The motivation of this part is to collectively analyze the group and its surroundings. This should also provide contextual information about the group i.e. where the group is and what type of event they are participating in. We use the Inception V3 \cite{szegedy2016rethinking} to train our model for predicting GCS. The main reason behind choosing inception V3 is that it provides a good trade-off between the number of parameters and accuracy in the case of the ImageNet challenge \cite{szegedy2016rethinking}. We have also conducted experiments on several deep Convolutional Neural Networks (CNNs), and results are shown in TABLE \ref{sota}. The inception V3 network is similar to the original work \cite{szegedy2016rethinking}, which was proposed for the classification on the ImageNet task except for the last few dense layers including the regression layer. The details of the layers are shown in TABLE \ref{img_arc}.

\begin{table}[t]
\centering
\begin{tabular}{|l|c|c|c|}
\hline
\textbf{Layers}                                              & \textbf{Input} & \textbf{Output} & \textbf{Layer Details} \\ \hline
Inception V3                                                        & b,224,224,3          & b,2048          & similar to \cite{szegedy2016rethinking}                     \\ \hline
Dense                                                        & b,2048          & b,4096          & 4096                     \\ \hline
Activation                                             & b,4096         & b,4096          & Relu/Swish             \\ \hline
Dense                                                        & b,4096         & b,4096          & 4096                     \\ \hline
Activation                                             & b,4096         & b,4096         & Relu/Swish             \\ \hline
Dense                                                        & b,4096         & b,4096          & 4096                     \\ \hline
Activation                                             & b,4096         & b,4096          & Relu/Swish             \\ \hline
\begin{tabular}[c]{@{}c@{}}Cohesion\\ (Sigmoid)\end{tabular} & b,4096           & b,1             & 1                     \\ \hline
\end{tabular} \vspace{ 2mm}
\caption{\small {Image-level network architecture. Here, \emph{b} and \emph{BN} refer to the batch size and batch normalization respectively.}}
\label{img_arc}
\vspace{-3mm}
\end{table}

In the word cloud of the survey result, people mentioned some group-level emotion-related keywords such as `violence', `happy', `angry', `upset' etc. Thus, we perform experiments with joint training for GCS and group emotion (three classes positive, neutral and negative \cite{dhall2017individual}). The motivation is to explore the usefulness of GCS of a group as an attribute for group emotion prediction. The network structure is the same as shown in TABLE \ref{img_arc} except for the last layer which predicts three group emotion probabilities and one GCS.

\begin{figure*}[!htbp]
    \centering
\subfloat[]{ \includegraphics[scale=0.4]{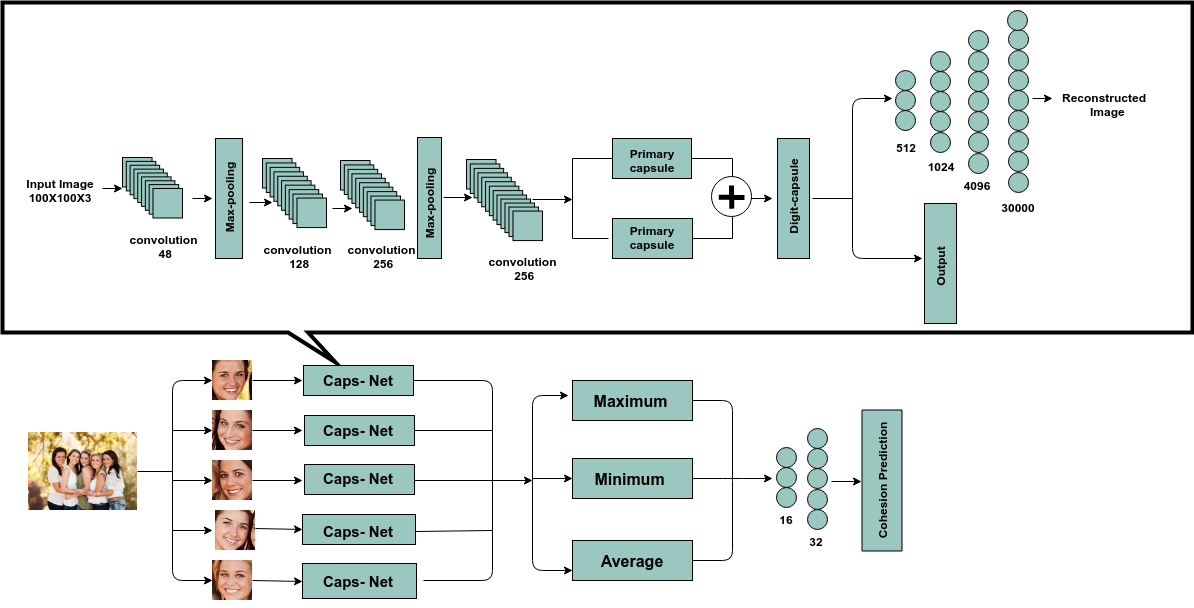}} \vspace{0.5mm}
\subfloat[]{ \includegraphics[width=\linewidth]{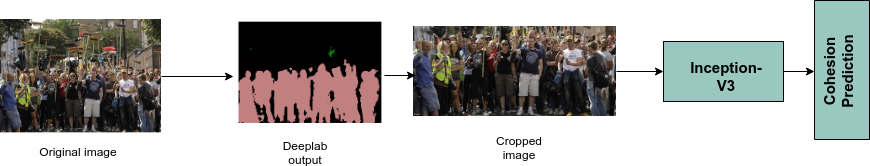}} \vspace{ 0.5mm}
% \subfloat[]{\includegraphics[width=\linewidth]{structure.png}} \vspace{ 2mm}
      \caption{\small  (a) CapsNet structure for face-level expression analysis. The prediction from this network is further pooled to predict the GCS. The face-level part first predicts expression (as shown in this figure) and then computes average, minimum and maximum. The details can be found in Section \ref{FaceLevel}.
      (b) Pipeline for the analysis of the background level importance using group segmentation. We crop the group \cite{deeplabv3plus2018} before inputting to the network for GCS prediction.
    %   (c) Network to encode structure information.
    }
      \label{cohesion_net_all}
      \vspace{-5mm}
\end{figure*}

\subsection{Face-Level Analysis} 
\label{FaceLevel}
Motivated by the result of joint training of the AGC and group emotion and survey results (apart from team, group keywords people are mainly focused on the mood like keywords such as angry, cheering, unhappy, violence etc.), we attempt to infer GCS based on the face-level emotion information as well. For facial emotion analysis, we use the recently proposed CapsNet \cite{sabour2017dynamic} architecture as shown in Fig. \ref{cohesion_net_all}. In order to overcome the drawbacks of traditional CNNs, Sabour et al. \cite{sabour2017dynamic} proposed a new CNN like architecture \textbf{Capsule Network} (CapsNet), which keeps the spatial orientation related information along with deep features. Here, capsules are a group of neurons which include the instantiation parameters of a certain object. For example, a face has eyes, nose, lips with certain constraints. The main difference between a CNN and a capsule Network is that the later stores the state of the feature (neuron output) in the form of a vector instead of a scalar. Another salient property of CapsNet is \textbf{routing by agreement}, which means activated capsules follow a hierarchy. Higher level capsules become activated if and only if lower level capsule outputs agree with it. As per \cite{sabour2017dynamic}, CapsNet is invariant to rotation and it can model a spatial hierarchy via \textbf{dynamic routing} and \textbf{reconstruction regularization}. Thus, the network can learn the pattern of viewpoint invariance between the object part and the whole object. From TABLE \ref{raf}, we can observe that CapsNet performs better than the other state-of-the-art networks. CapsNet can explicitly model the pose and illumination of an object. Inspired by this argument, we choose to train CapsNet.
\begin{table}[t]
\centering
\scalebox{0.9}{
\begin{tabular}{|c|c|c|c|c|}
\hline
\textbf{RAF-DB}                                                     & \textbf{Ours} & \textbf{Alexnet} & \textbf{mSVM \cite{li2017reliable}} & \textbf{DLPCNN \cite{li2017reliable}} \\ \hline
\begin{tabular}[c]{@{}c@{}} Accuracy(\%)\end{tabular} & 77.48         & 76.27      & 65.12         & 74.20           \\ \hline
\end{tabular}
% \begin{tabular}{|c|c|c|c|c|}
% \hline
% \textbf{Network} & \textbf{DLP CNN} & \textbf{Alexnet} & \textbf{Inception V3} & \textbf{\begin{tabular}[c]{@{}c@{}}Capsule \\ Network\end{tabular}} \\ \hline
% Accuracy(\%)     & 76.27            & 73.00            & 74.20                 & 77.48                    \\ \hline
% \end{tabular}
} 
\caption{\small Comparison of the performance of CapsNet with other networks on RAF-DB.}
\label{raf}
\vspace{-2mm}
\end{table}
We slightly modified the proposed architecture of CapsNet \cite{sabour2017dynamic} used for digit classification. 
%The main motivation behind this network is to copy the human visual system which regenerates the view of the image it observes. 
CapsNet takes cropped face as input and predicts the seven basic emotions (i.e. \emph{happy}, \emph{neutral}, \emph{sad}, \emph{angry}, \emph{surprise}, \emph{disgust} and \emph{fear}) as a output. Thus, we get emotion probability predictions for each of the faces present in a group image. Further, we pool the predicted emotion labels by computing the average, maximum and minimum (get $batch size\times 3 \times 7 $ dimensional output). This small feature is then fed to two dense layers of 16 and 32 nodes (empirically) respectively before predicting the GCS. This whole network structure is shown in TABLE \ref{face_arc}.

\begin{table}[t]
\centering
\begin{tabular}{|c|c|c|c|}
\hline
\textbf{Layers}                                              & \textbf{Input} & \textbf{Output} & \textbf{Layer Details} \\ \hline
Dense                                                        & b,3,7          & b,3,16          & 16                     \\ \hline
BN and Activation                                             & b,3,16         & b,3,16          & Relu/Swish             \\ \hline
Dense                                                        & b,3,16         & b,3,32          & 32                     \\ \hline
BN and Activation                                             & b,3,32         & b,3,32          & Relu/Swish             \\ \hline
Max Pooling                                                  & b,3,32         & b,1,32          & 3(1-D)                 \\ \hline
Flatten                                                      & b,1,32         & b,32            &  -                      \\ \hline
\begin{tabular}[c]{@{}c@{}}Cohesion\\ (Sigmoid)\end{tabular} & b,32           & b,1             & 1                     \\ \hline
\end{tabular}
\caption{\small Face-level Network Architecture. Here, b and BN refer to the batch size and batch normalization respectively.}
\label{face_arc}
\vspace{-5mm}
\end{table}

% \begin{figure}[h]
%     \centering
%       \includegraphics[width=\linewidth]{deeplab.png}
%       \caption{ Pipeline for the analysis of the background level importance using group segmentation. We crop the group \cite{deeplabv3plus2018} before inputing to the network for GCS prediction.}
%       \label{deeplab}      
% \end{figure}

\subsection{Effect of Background}
Further, we also investigate how the background effects AGC. We use the segmentation technique to crop people from group images via Deeplab V3plus \cite{deeplabv3plus2018} library. We consider an area-wise threshold, i.e. if the segmented area is less than 50\% of the total area of the image then this image is considered for analysis (pipeline is shown in Fig. \ref{cohesion_net_all}). Thus, we observed that when we use the segmented image for training, then there is a drop (around 0.103 MSE decreased) in performance. It indicates that the background around a person also plays a vital role in the perception of a group's cohesiveness. The background may reflect something about the social event in which the group is participating and is important for the prediction.

% \subsection{Effect of Group Structure}
% In order to encode group structure information, we follow the following steps:
% \begin{itemize}
% \item Detect faces via MTCNN \cite{zhang2016joint} and takes the tip of the nose as face location.
% \item Perform k-means clustering of these points and find the centroid of the group.

% \item Detect the boundary of the group structure as shown in Fig. \ref{cohesion_net_all}. (we use single region alpha shape \cite{edelsbrunner1983shape} with 0.5 shrink factor for this purpose)

% \item Divide the angle around the centroid into 64 folds. Thus,
% $$ \theta _{j} = \dfrac{2 \pi} {64} * j, where j \in \{ 1, 2..., 64 \}. $$

% \item From the centroid to this boundary, we take 64 radii (R) of this polygon $ R= L(\theta _{j}) $. Here, R is a vector of length=64.

% \item Perform Fast Fourier Transform (FFT) of the R and extract the feature along its amplitude spectrum. For simplicity, we took 100-dimensional feature vector for this purpose.

% \item Further, with the above-mentioned feature, we train a small DNN of FC layers (64, 32, 1) for cohesion prediction.  
% \end{itemize}

% \subsection{Merged Network}

% For predicting overall AGC on the basis of structure, emotion and holistic features, we take second last layer output, concatenate it and pass it through several dense layers (1024, 512, 128) to extract overall AGC score.

% \begin{figure}[!htbp]
%     \centering
%       \includegraphics[width=\linewidth]{structure.png}
%       \caption{ Network to encode structure information.}
%       \label{structure}
      
% \end{figure}

\section{Experimental Details and Results}
\label{exp}
In this section, we discuss the experimental settings and results. First of all, we treat cohesion as a regression problem (as also defined in \cite{treadwell2001group}) and the group emotion as a classification problem (defined in \cite{dhall2017individual}). We use the Keras \cite{chollet2015keras} deep learning library for the implementation. 

\subsection{Image-Level Analysis Results}
We train Inception V3 network for image-level analysis. We initialize the network with ImageNet pre-trained weights and fine-tune the network with SGD optimizer having a learning rate of 0.001 and momentum 0.9 without any learning rate decay. Our image-level experimental results are shown in TABLE \ref{overall}. With similar hyperparameters, we jointly train an inception V3 network for both emotion and cohesion prediction. The results (TABLE \ref{image}) show an interesting pattern. When the inception V3 is individually used for group emotion and cohesion prediction, its performance is lower than the joint training. Thus, it suggests that the network learns more relevant representations of group emotion. We can conclude that the emotion and cohesion at group-level are interrelated terms. Human perception behind group emotion and cohesion has some sort of similarity.
\begin{table}[b]
\centering
\begin{tabular}{|c|c|c|}
\hline
\textbf{Network}                                                                                      & \textbf{Accuracy (\%)} & \textbf{MSE} \\ \hline
\begin{tabular}[c]{@{}c@{}}Inception V3 \\(emotion and cohesion prediction)\end{tabular} & 85.58              &0.8181              \\ \hline
\begin{tabular}[c]{@{}c@{}}Inception V3 (emotion prediction)\end{tabular}                 & 65.41              & NA              \\ \hline
\begin{tabular}[c]{@{}c@{}}Inception V3 (cohesion  prediction)\end{tabular}                &   NA
& 0.8537             \\ \hline
\end{tabular} \vspace{ 2mm}
\caption{\small The results of image-level group emotion (classification accuracy) and cohesion (MSE) analysis. \emph{Notice that the performance of group emotion increases, when it is jointly trained with AGC. However, the same is not true for AGC}.}
\label{image}
\vspace{-5mm}
\end{table}
This is in accord with psychology studies \cite{barsade1998group}. It is also interesting to note that the effect of joint training is opposite to the GCS prediction as the prediction error increases. One possible reason is that GCS and group emotion features contradict each other. Let us consider the example of a sobbing family, which has high GCS and negative group emotion and compare that with of a celebrating sports team, which will also have high GCS. In the later, the group emotion will be positive. Scenarios like this may lead to ambiguity during the joint training from the GCS prediction perspective.

\begin{table}[b]
\centering \vspace{-1mm}
\begin{tabular}{|c|c|c|c|c|}
\hline
\textbf{Network Details}  & \textbf{Image-Level} & \textbf{Face-level} & \textbf{EmotiW Baseline}\\ \hline
GCS (MSE on Val. set)                      & 0.85                  & 1.11   & 0.84         \\ \hline
GCS (MSE on Test set)                      & 0.53                  & 0.91     &0.50        \\ \hline
\end{tabular}

\caption{\small Comparisons of GCS prediction using the image-level and face-level networks. Due to copyright we will release a subset of the data for the EmotiW 2019 challenge. The baselines are mentioned in the last column of this table.}
\label{overall}
\vspace{-3mm}
\end{table}

% \begin{table}[!htbp]
% \centering \vspace{ 1mm}
% \begin{tabular}{|c|c|c|c|c|c|}
% \hline
% \textbf{Network Details}  & \textbf{Image-Level} & \textbf{Face-level} & \textbf{Group Structure} & \textbf{Merged}\\ \hline
% GCS (MSE on Val. set)                      & 0.853                  & 1.113         & 0.758   & 0.751\\ \hline
% GCS (MSE on Test set)                      & 0.532                  & 0.914          & 0.497  & 0.516 \\ \hline
% \end{tabular}
% \caption{Comparison of GCS prediction using described networks.}
% \label{overall}
% \end{table}

\begin{table}[!htbp]
\centering 
\begin{tabular}{|c|c|c|}
\hline
\textbf{Cross validation} & \textbf{\begin{tabular}[c]{@{}c@{}}MSE \\ (lr=0.001)\end{tabular}} & \textbf{\begin{tabular}[c]{@{}c@{}}MSE \\ (lr=0.01)\end{tabular}} \\ \hline
1$^{st}$                       & 0.63958                                                            & 0.65662                                                           \\ \hline
2$^{nd}$                       & 1.10628                                                            & 1.06666                                                           \\ \hline
3$^{rd}$                       & 0.70162                                                            & 0.67964                                                           \\ \hline
4$^{th}$                       & 0.60604                                                            & 0.76320                                                            \\ \hline
5$^{th}$                       & 0.93969                                                            & 0.89159                                                           \\ \hline
\textbf{Average}              & \textbf{0.79864}                                                   & \textbf{0.81155}                                                  \\ \hline
\end{tabular}
\caption{\small 5 fold cross validation results of the GAF cohesion database. lr = learning rate}
\label{crossval}
\vspace{-3mm}
\end{table}

\begin{figure*}[t]
    \centering
      \includegraphics[width=\linewidth]{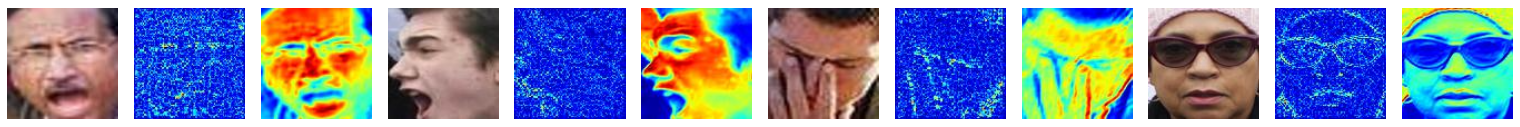}
      \caption{\small Visualization of facial emotion. Each set of three images shows the original image, saliency map and class activation map respectively. It is the class activation map of the CapsNet trained for emotion task. Here, red coloured region indicates activated regions.  It is visible that the CapsNet can handle non-frontal, occluded, scaled and rotated images properly. Statistics (Maximum, Minimum and Average) of these features further used for predicting overall AGC. [Best viewed in colour]}
\label{vis_face}
\vspace{-5mm}
\end{figure*}

\subsection{Face-Level Analysis Results}
In order to predict GCS, we pre-trained a CapsNet (Capsule Net) on RAF-DB \cite{li2017reliable}. RAF-DB  \cite{li2017reliable} is a facial expression database containing around 30K single person face images which are labeled for seven basic emotion classes as well as 12 compound emotions. We take basic emotions (i.e. \emph{happy}, \emph{neutral}, \emph{sad}, \emph{angry}, \emph{surprise}, \emph{disgust} and \emph{fear}) to train our CapsNet. From group images, we extracted faces via MTCNN \cite{zhang2016joint}. After training on RAF-DB, we take the output probability vector for each face in the group image. Further, we compute statistics over these emotion probabilities and pass it through two more dense layers before final cohesion score prediction. Our statistics include maximum, average and minimum of respective emotion probabilities. The motivation behind this is that we need to conclude over a group. Hence, these three intensity level analyses perform better for group-level tasks. 

We train a CapsNet with hyperparameters from the original paper (Adam optimizer with default settings in the Keras library and learning rate decay of 0.001 in every 10$^{th}$ epoch to avoid local minima). We train the rest of the network via SGD optimizer with learning rate 0.01 and without any learning rate decay.
TABLE \ref{overall} presents the results of the image-level and face-level networks in GAF Cohesion database. TABLE \ref{crossval} describes 5-fold cross-validation results of the GAF cohesion database. In TABLE \ref{sotagaf}, we predict group emotion, when AGC information is used for joint training. The results of the GAF 2.0 show better performance than the other state-of-the-art methods. This shows that cohesiveness information is useful for group emotion prediction. 

\begin{table}[t]
\centering
\begin{tabular}{|c|c|c|c|c|c|}
\hline
\textbf{GAF 2.0} & \textbf{Ours} & \textbf{\cite{tan2017group}} & \textbf{\cite{guo2017group}} & \textbf{\cite{wei2017new}} & \textbf{\cite{dhall2017individual}} \\ \hline
Accuracy(\%)     & 85.67         & 83.90             & 80.05            & 77.92            & 52.97             \\ \hline
\end{tabular}
\caption{\small Group emotion performance comparison on GAF.}
\label{sotagaf}
\vspace{-3mm}
\end{table}

\section{Visualization (Saliency vs Class Activation)}
\label{sec:visual}
\begin{figure}[t]
\centering 
\subfloat{\includegraphics[width = 1in,height=1in]{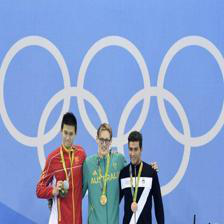}}
\subfloat{\includegraphics[width = 1in,height=1in]{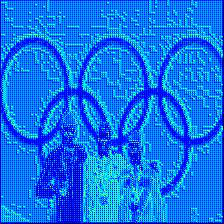}}
\subfloat{\includegraphics[width = 1in,height=1in]{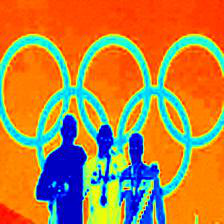}}\\\vspace{-4mm}
\subfloat{\includegraphics[width = 1in,height=1in]{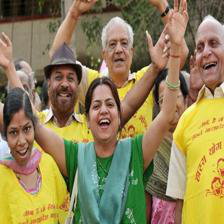}}
\subfloat{\includegraphics[width = 1in,height=1in]{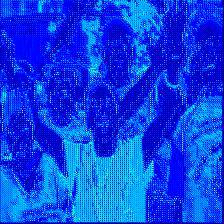}}
\subfloat{\includegraphics[width = 1in,height=1in]{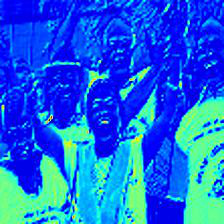}}\\\vspace{-4mm}
\subfloat{\includegraphics[width = 1in,height=1in]{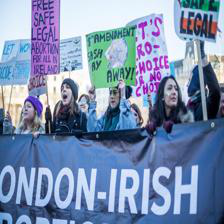}}
\subfloat{\includegraphics[width = 1in,height=1in]{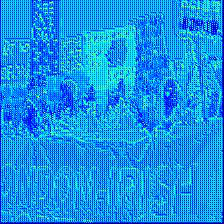}}
\subfloat{\includegraphics[width = 1in,height=1in]{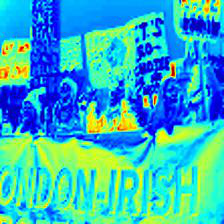}}\\\vspace{-4mm}
\subfloat{\includegraphics[width = 1in,height=1in]{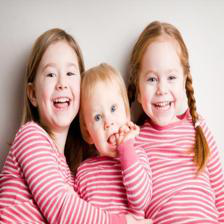}}
\subfloat{\includegraphics[width = 1in,height=1in]{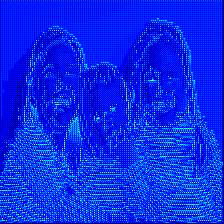}}
\subfloat{\includegraphics[width = 1in,height=1in]{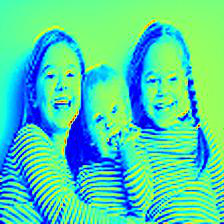}} \vspace{-2mm}
\caption{\small Visualization of image-level cohesion. Each row consists of the original image, saliency map and class activation map. The top row focuses on background features, the second row focuses on foreground features, the third row focuses on context level feature and the last row focuses on facial regions. [Best viewed in colour]}
\label{vis_img}
\vspace{-5mm}
\end{figure}

In this section, we discuss visual attributes that our network learns. We visualize the class activation map and discuss its comparison with the saliency. 
% For visualization purpose, we use the Keras vis library. 
From Fig. \ref{vis_face}, we can observe that in spite of non-frontal, rotated, occlusion, blurred faces, CapsNet can handle each case efficiently. Especially, it deals with the rotation and scaling of different objects in an image individually and shows better performance over both occluded and partially occluded images which is beneficial for our problem. Moreover, it did not require data augmentation while training and thus it is efficient regarding time complexity. Similarly, for image-level analysis, (as shown in Fig. \ref{vis_img}) the top row activates the background, the second row activates the foreground, the third row activates the subject and the last row activates both the front person and background. In the case of the top row, it activates the background, as the group takes up a small space as compared to the visible background. Similarly, in the second case the foreground is more dominant as compared to the background. In the third row, the main features of the protests that are activated are the banners. In the last picture, it activates both foreground and background, especially the facial region.
The image saliency and class activation have a significant difference. From Fig. \ref{vis_face} and Fig. \ref{vis_img}, respectively, it is visible that AGC is not directly labeled as well as predicted via image saliency. Although, there are some regions in images, which are common in both cases. Of course, sometimes the human mind is also influenced by some salient feature of the images.

\section{Conclusion, Limitation and Future Work}
\label{con}
The main motivation behind our approach is to achieve human-level perception regarding cohesion and to study the reason behind it. Here, we use the term human-level because we consider human annotation as the ground truth. 
%It is evident from the results that we have achieved near human level performance\footnote{Here, we use the term human level because we consider human annotation as ground truth.} in AGC task. Our model performs comparatively well and achieves near the human level performance.
From our experimental results, it can be deduced that AGC and emotion are interrelated. This work leverages the usefulness of both facial and scene features. We also observed that the newly proposed CapsNet \cite{sabour2017dynamic} also performs well on facial expression recognition without data augmentation. Although the faces in a group image vary largely, i.e. the face can be occluded, blurred or non-frontal and others. Via visualization, we observed that scene information encodes the background, clothes and various objects in an image. This information is also known as the top-down contextual information. The main limitation of our work is the cultural influence in data annotation as it is related to the perception of cohesion. A potential future direction for our work is to investigate how facial attributes effect AGC prediction. It will be interesting to analyze the role of the body pose of the group members along with the face. Although, the image-level network does encode the body pose, however, its complete contribution requires further investigation. It will be of interest to analyze the fashion quotient of the group by parsing the clothes for patterns and themes, which correspond to specific social events, although, some patterns are already encoded in our scene level analysis. Furthermore, another possible direction is to include kinship related information in the network because irrespective of visual expression, sometimes kinship indicates strong cohesion.

\section*{Acknowledgement}
We acknowledge the support of NVIDIA for providing us TITAN Xp G5X GPU for research purposes. 
%We would also like to show our gratitude to our lab mates for sharing their pearls of wisdom with us during the course of this research. 
% ``Fig.~\ref{fig}'', even at the beginning of a sentence.

% \begin{table}[htbp]
% \caption{Table Type Styles}
% \begin{center}
% \begin{tabular}{|c|c|c|c|}
% \hline
% \textbf{Table}&\multicolumn{3}{|c|}{\textbf{Table Column Head}} \\
% \cline{2-4} 
% \textbf{Head} & \textbf{\textit{Table column subhead}}& \textbf{\textit{Subhead}}& \textbf{\textit{Subhead}} \\
% \hline
% copy& More table copy$^{\mathrm{a}}$& &  \\
% \hline
% \multicolumn{4}{l}{$^{\mathrm{a}}$Sample of a Table footnote.}
% \end{tabular}
% \label{tab1}
% \end{center}
% \end{table}

% \begin{figure}[htbp]
% \centerline{\includegraphics{fig1.png}}
% \caption{Example of a figure caption.}
% \label{fig}
% \end{figure}
{\small 
\bibliographystyle{IEEEtran}  
\bibliography{IEEEabrv,ieee}
}
% \bibitem{b1} G. Eason, B. Noble and I. N. Sneddon, ``On certain integrals of Lipschitz-Hankel type involving products of Bessel functions,'' Phil. Trans. Roy. Soc. London, vol. A247, pp. 529--551, April 1955.
% \bibitem{b2} J. Clerk Maxwell, A Treatise on Electricity and Magnetism, 3rd ed., vol. 2. Oxford: Clarendon, 1892, pp.68--73.
% \bibitem{b3} I. S. Jacobs and C. P. Bean, ``Fine particles, thin films and exchange anisotropy,'' in Magnetism, vol. III, G. T. Rado and H. Suhl, Eds. New York: Academic, 1963, pp. 271--350.
% \bibitem{b4} K. Elissa, ``Title of paper if known,'' unpublished.
% \bibitem{b5} R. Nicole, ``Title of paper with only first word capitalized,'' J. Name Stand. Abbrev., in press.
% \bibitem{b6} Y. Yorozu, M. Hirano, K. Oka and Y. Tagawa, ``Electron spectroscopy studies on magneto-optical media and plastic substrate interface,'' IEEE Transl. J. Magn. Japan, vol. 2, pp. 740--741, August 1987 [Digests 9th Annual Conf. Magnetics Japan, p. 301, 1982].
% \bibitem{b7} M. Young, The Technical Writer's Handbook. Mill Valley, CA: University Science, 1989.
% \end{thebibliography}

\end{document}